# SEDD-PCC: A SINGLE ENCODER – DUAL DECODER FRAMEWORK FOR END-TO-END LEARNED POINT CLOUD COMPRESSION

*Kai-Hsiang Hsieh, Monyneath Yim, Jui-Chiu Chiang*

Department of Electrical Engineering, National Chung Cheng University, Taiwan

## ABSTRACT

To encode point clouds containing both geometry and attributes, most learning-based compression schemes treat geometry and attribute coding separately, employing distinct encoders and decoders. This not only increases computational complexity but also fails to fully exploit shared features between geometry and attributes. To address this limitation, we propose SEDD-PCC, an end-to-end learning-based framework for lossy point cloud compression that jointly compresses geometry and attributes. SEDD-PCC employs a single encoder to extract shared geometric and attribute features into a unified latent space, followed by dual specialized decoders that sequentially reconstruct geometry and attributes. Additionally, we incorporate knowledge distillation to enhance feature representation learning from a teacher model, further improving coding efficiency. With its simple yet effective design, SEDD-PCC provides an efficient and practical solution for point cloud compression. Comparative evaluations against both rule-based and learning-based methods demonstrate its competitive performance, highlighting SEDD-PCC as a promising AI-driven compression approach.

***Index Terms*—** Point Cloud Compression, Unified Model, Geometry Compression, Attribute Compression, knowledge distillation

## 1. INTRODUCTION

Point clouds are a widely used 3D data representation, consisting of millions of unordered points defined by their geometric coordinates and attributes such as color, reflectance, and normals. Their ability to accurately capture the geometry and appearance of objects and scenes makes them ideal for immersive multimedia applications, including augmented and virtual reality (AR/VR), autonomous driving, gaming, robotics, and cultural heritage preservation. However, their inherently large size demands efficient compression techniques for practical storage, transmission, and real-time streaming. To address the challenges of point cloud compression, the Moving Picture Experts Group (MPEG) developed two standards: Video-Based Point Cloud Compression (V-PCC) [1] and Geometry-Based Point Cloud Compression (G-PCC) [2]. These standards efficiently

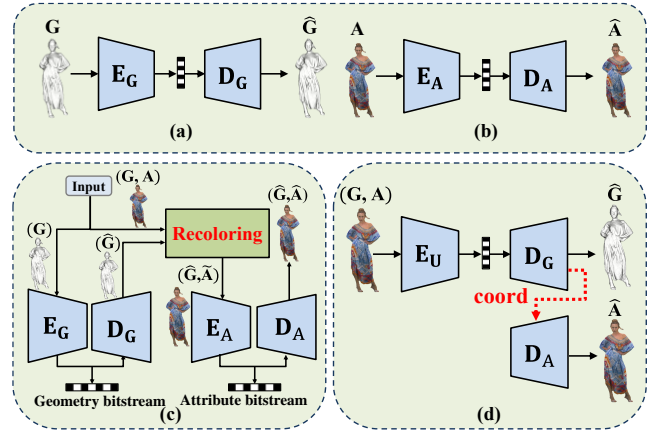

Fig. 1 Existing Learning-based point cloud compression solutions, (a) Geometry compression. (b) Attribute compression. (c) Joint geometry and attribute compression through recoloring (d) proposed SEDD-PCC: an end-to-end learned point cloud compression scheme.

handle both static and dynamic point clouds, balancing compression efficiency and data fidelity.

Recently, learning-based approaches have shown promising results in point cloud compression. Recognizing this potential, MPEG and JPEG have initiated efforts toward standardizing AI-driven compression through the AI-PCC framework. Many existing works focus on geometry compression [3-7] (Fig. 1a), while more recent efforts have explored attribute compression [8-13] (Fig. 1b), often assuming a known geometry. Besides, some works [14-15] attempt to jointly compress both geometry and attributes, with the first fully learned model [14] representing point clouds as a 4-channel 3D voxel grid where one channel encodes occupancy and the remaining three store color information. However, the model in [14] often lack specialized design, leading to suboptimal performance compared to traditional approaches such as G-PCC. The first version of the JPEG Pleno PCC verification model [15], similar to [14], adopts the same input representation but features with advanced networks and integrates super-resolution modules, outperforming G-PCC.

Other joint compression solutions [16-19, 21] adopts a sequential pipeline where geometry is first encoded and decoded, followed by attribute recoloring and a separate encoding-decoding stage for attribute (Fig. 1c). Notable works include: YOGA [16] which employs a variable-rate design with G-PCC as the base layer and a neural network for enhancement; DeepPCC [17] leveraging sparse convolution and local self-attention to effectively model spatial relationships, improving the accuracy of geometry occupancy probability and attribute intensity estimation; Unicorn [20, 21] introduces a multiscale conditional coding framework that supports lossy and lossless compression, static and dynamic coding to cover various point cloud types, marking a significant step forward in AI-PCC; Meanwhile, building upon the JPEG Pleno PCC standard [18], the latest update, JPEG Pleno PCCv4.1 [19] further advances AI-driven attribute coding by integrating JPEG AI [22], achieving compression performance well beyond that of G-PCC.

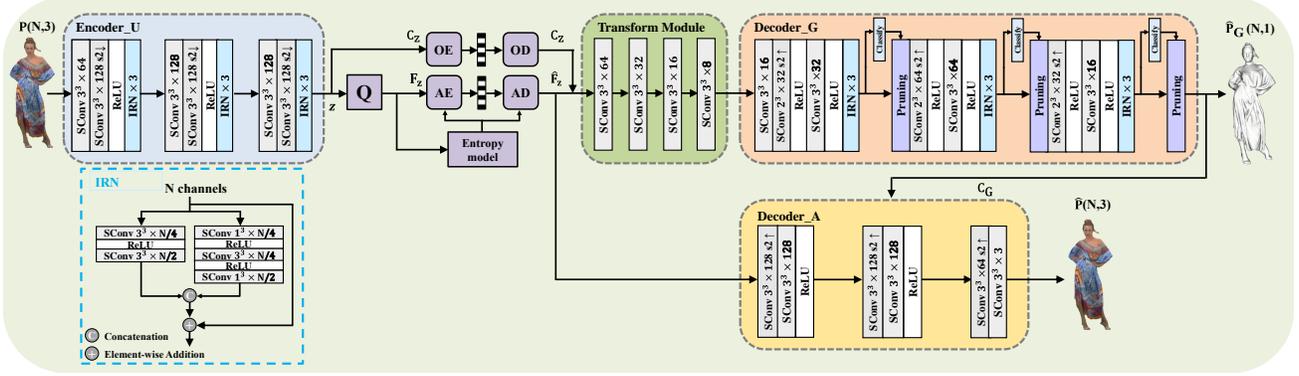

Fig. 2 The proposed SEDD-PCC architecture

Currently, sequential approaches outperform unified models that use a single encoder and decoder. However, they present several challenges. One major issue is error propagation, where distortions in the reconstructed geometry are transferred to the attributes during the recoloring process, ultimately degrading attribute quality. Additionally, encoding geometry and attributes separately prevents the model from jointly optimizing the feature space, leading to suboptimal compression efficiency. The use of two distinct codecs along with a recoloring module further increases computational complexity, requiring additional processing steps and memory overhead. Moreover, bit allocation between geometry and attribute coding remains a nontrivial challenge, making it difficult to achieve an optimal balance for overall reconstruction quality.

To address these limitations, we propose SEDD-PCC, a novel unified model that employs a single encoder to jointly learn features for both geometry and attributes, leveraging their strong correlations. The model reconstructs geometry and attributes sequentially using two dedicated decoders, as illustrated in Fig. 1(d). While this design offers significant advantages, it also introduces challenges. One major issue is the conflicting optimization objectives, as the encoder must learn a shared latent space that serves both decoders. This may lead to interference, where features essential for geometry reconstruction are influenced by attribute variations, ultimately degrading overall quality. Furthermore, the shared latent space may struggle to generalize effectively, particularly when geometry and attributes require different levels of detail.

To overcome these challenges, we introduce a multi-stage training strategy and incorporate knowledge distillation to enhance geometry reconstruction. Specifically, the training process consists of three stages: first, the shared encoder is trained alongside the attribute decoder to ensure effective attribute representation learning. Next, the geometry decoder is trained separately to refine geometric reconstructions without being affected by attribute variations. Finally, a fine-tuning stage optimizes the entire model to ensure both decoders work synergistically, improving overall compression performance. Our contributions include:

- We propose SEDD-PCC, the first unified model that employs a single encoder with two specialized decoders for joint compression of point cloud geometry and attributes, leveraging a simple yet effective autoencoder architecture.
- We introduce a three-stage training strategy, where the shared encoder and attribute decoder are trained first, followed by geometry decoder, and a final fine-tuning stage to enhance overall compression performance.
- SEDD-PCC achieves a Bjøntegaard Delta bitrate (BD-BR) reduction of 75.0% for D1-PSNR, 32.6% for Y-PSNR, and 33.2% for 1-PCQM metrics compared to G-PCC. Additionally, it delivers competitive performance against other learning-based methods while maintaining a lightweight and efficient design, demonstrating its potential as a robust solution for AI-driven point cloud compression.

## 2. PROPOSED METHOD

### 2.1 SEDD-PCC

The proposed SEDD-PCC architecture, illustrated in Fig. 2, consists of a single encoder and dual decoders designed for joint compressing of geometry and attributes. Since point cloud attributes are inherently tied to their corresponding geometry, attributes cannot exist independently. Leveraging this nature, our shared encoder processes the point cloud input as a three-channel 3D voxel grid. The input point cloud is represented as a sparse tensor with coordinates $C = \{(x_i, y_i, z_i) \mid i \in [0, N-1]\}$ and features $\mathbf{F} = \{(R, G, B) \mid i \in [0, N-1]\}$, where $N$ is the total number of points.

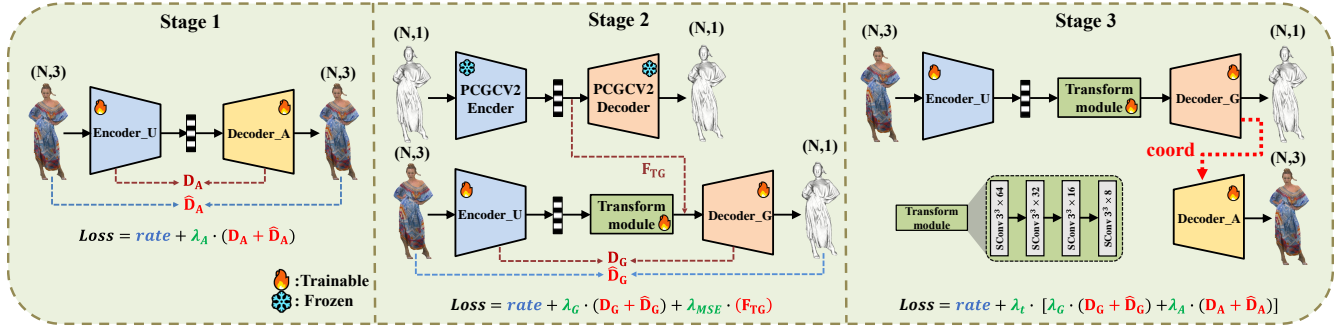

Fig. 3 The three-stage training strategy. Stage 1: attribute compression; Stage 2: geometry compression, adopting a teacher model for knowledge distillation; Stage 3: Fine-tune all components by initializing Encoder_U and Decoder_A from Stage 1, and Decoder_G from Stage 2.

Unlike previous approaches [14, 15] that use a single encoder-decoder pair and represent the input as a four-channel tensor, our design adopts a shared encoder that processes only three channels, focusing exclusively on attributes. This choice allows for a more compact representation of point clouds, as attributes inherently encapsulate richer information than the geometry occupancy. In contrast, four-channel representations may introduce less relevant joint features, potentially limiting compression efficiency.

The primary challenge in our design lies in constructing an effective shared encoder. Since attribute compression is inherently more complex than geometry compression, we propose that the shared encoder should be structured more like an attribute encoder to facilitate richer feature extraction rather than a geometry-centric approach. To achieve this, we build upon Sparse-PCAC [9], integrating Inception-Residual Network (IRN) layers [23] in both the encoder and decoder, enhancing feature extraction and improving reconstruction quality. Once the shared encoder is designed as an attribute encoder, the next challenge is constructing the geometry decoder to reconstruct features from the shared latent space. Because the shared encoder outputs more channels than a typical geometry encoder, a transform module is introduced to convert these features effectively.

### 2.2 Encoding and Decoding

The encoding process begins by voxelizing the input point cloud into a structured sparse tensor with three channels. This voxelized data is then passed through the shared encoder, where sparse convolutions progressively transform it into a compact latent representation $z$. The thumbnail point cloud geometry $C_z$ is losslessly encoded using the G-PCC octree codec, while the corresponding feature $F_z$ is quantized and entropy encoded. During decoding, the quantized feature $F_z$ is first processed by the transform module before being passed to the geometry decoder, which reconstructs the point cloud structure. The decoder follows a similar hierarchical approach as the encoder, progressively upscaling and classifying points using the Top-$k$+1 mechanism [24]. Once the geometry is reconstructed, attribute decoding follows the Sparse-PCAC [9] process to complete point cloud reconstruction.

### 2.3 Training Protocol

Using a single encoder with two decoders for geometry and attributes introduces challenges due to conflicting optimization objectives. The encoder must learn a shared latent space that serves both decoders, but geometry features may be influenced by attribute variations, potentially degrading reconstruction quality. Moreover, this setup may limit generalization, as it must balance different levels of detail required for geometry and attributes. To address these challenges, we adopt a three-stage training approach, as illustrated in Fig. 3. This method consists of an attribute coding stage, a geometry coding stage, and a joint coding stage, each with a dedicated loss function using a Lagrangian loss: $R + \lambda D$, where $R$ denotes the bit rate, $D$ represents the distortion, and $\lambda$ serves as the trade-off parameter to balance the two terms.

In the first stage, attribute coding is trained, assuming the geometry is known, and the loss function is shown in (1), computing the distortion in the YUV color space. The first term $d(x, \hat{x})$, represents the mean squared error (MSE) between the original and the reconstructed point cloud. The second term $d(x, \bar{x})$, denotes the MSE of the multi-scale loss between the encoder and decoder, aimed at enhancing reconstruction quality.

$$\mathcal{L}_A = R + \lambda_A \cdot D = R_{\hat{z}} + \lambda_A \cdot (\alpha \cdot \|x - \hat{x}\|_2^2 + \|x - \bar{x}\|_2^2), \quad (1)$$

The second stage focuses on geometry coding. Since the shared encoder is designed similarly to attribute encoders, in addition to the usage of a transform module, we also use PCGCV2 [3] as a teacher model for knowledge distillation. Our decoder is similar to PCGCv2, and we utilize a transform module to perform knowledge distillation, enabling learned features to resemble those of the teacher model. The loss function of this stage is shown in (2), which consist of three terms. The first term $\mathcal{L}_{BCE}$, represents binary cross-entropy (BCE) loss between the original and the reconstructed point cloud. The second term $\mathcal{L}_{BCE2}$, denotes the multi-scale BCE loss between the encoder and decoder. The third term $d(F_S, F_T)$, represents the MSE loss between feature representations from the student and teacher models.

Table 1. BD-rate (%) against G-PCCv23 for various schemes.

| Sequence | V-PCCv22 | | | YOGA [16] | | | DeepPCC [17] | | | JPEG Pleno [19] | | | Proposed (SEDD-PCC) | | |
|---|---|---|---|---|---|---|---|---|---|---|---|---|---|---|---|
| | D1 | Y | 1-PCQM | D1 | Y | 1-PCQM | D1 | Y | 1-PCQM | D1 | Y | 1-PCQM | D1 | Y | 1-PCQM |
| Longdress | -66.8 | -64.7 | -50.1 | -86.1 | -48.5 | -32.4 | -83.7 | -49.8 | -47.5 | --- | --- | --- | -71.4 | -34.5 | -35.1 |
| Loot | -74.4 | -72.2 | -64.9 | -87.0 | -58.9 | -53.2 | -84.9 | -51.0 | -55.4 | --- | --- | --- | -71.7 | -42.7 | -43.8 |
| Redandblack | -66.1 | -64.7 | -52.2 | -84.9 | -55.0 | -48.5 | -81.3 | -53.2 | -54.3 | --- | --- | --- | -66.2 | -42.0 | -35.3 |
| Soldier | -64.0 | -60.5 | -53.1 | -85.3 | -53.2 | -46.7 | -81.8 | -54.5 | -61.1 | -62.9 | -51.5 | -42.3 | -71.7 | -50.3 | -53.9 |
| Basketball player | -85.3 | -66.2 | -49.7 | -60.0 | 38.3 | 89.4 | -93.0 | -17.5 | 11.5 | --- | --- | --- | -85.3 | -13.0 | -14.7 |
| Dancer | -83.6 | -66.5 | -46.4 | -55.5 | 47.0 | 106.0 | -86.2 | -6.24 | 12.7 | --- | --- | --- | -83.8 | -12.7 | -16.1 |
| **Average** | **-73.4** | **-65.8** | **-52.7** | **-76.5** | **-21.2** | **2.44** | **-86.2** | **-38.7** | **-32.4** | **-62.9** | **-51.5** | **-42.3** | **-75.0** | **-32.6** | **-33.2** |

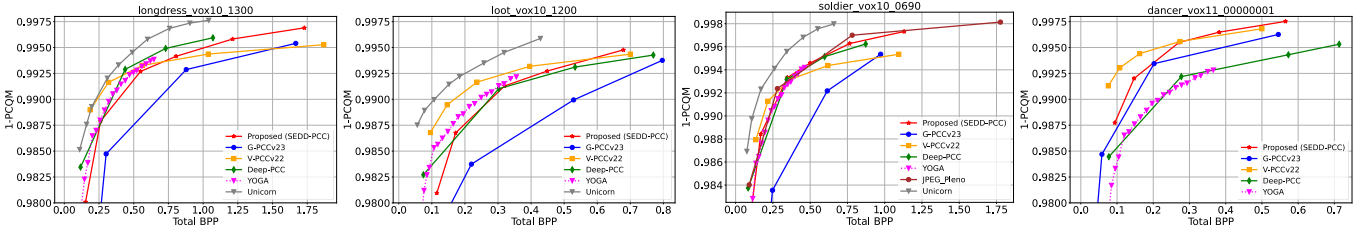

Fig. 4 R-D performance of the proposed scheme in terms of 1-PCQM.

$$\mathcal{L}_G = R + \lambda_G \cdot D + \lambda_{MSE} \cdot D_{feature}$$
$$= R_{\hat{z}} + \lambda_G \cdot (\mathcal{L}_{BCE} + \mathcal{L}_{BCE2}) + \lambda_{MSE} \cdot (\|F_S - F_T\|_2^2), \quad (2)$$

The final stage fine-tunes all components. The shared encoder initializes with weights from stage 1, while each decoder inherits weights from its respective previous stage. The loss function, as shown in (3), optimizes the overall network. Attribute MSE is computed in both forward and backward directions: the first computes MSE by mapping each point in the original point cloud to its closest point in the reconstructed cloud, while the second reverses this process. The overall distortion measure takes the maximum value between these two. This multi-stage approach ensures robust reconstruction and improved performance, effectively balancing bit rate and distortion while leveraging knowledge distillation to enhance feature representation.

$$\mathcal{L}_U = R + \lambda_t \cdot (\lambda_A \cdot D_A + \lambda_G \cdot D_G)$$
$$= R_{\hat{z}} + \lambda_t \cdot [\lambda_A \cdot (\alpha \cdot \|x - \hat{x}\|_2^2 + \|x - \bar{x}\|_2^2) + \lambda_G \cdot (\mathcal{L}_{BCE} + \mathcal{L}_{BCE2})], \quad (3)$$

## 3. EXPERIMENTS
### 3.1 Experimental Setting

***Training Datasets.*** For our training dataset, we select ScanNet [25], which contains over 1,500 highly detailed 3D indoor scenes. To handle GPU memory constraints during training, we divide the original point cloud data into non-overlapping 6-bit-sized cubes in each dimension. From these partitions, 50,000 cubes are used for training. The implementation is carried out using PyTorch and MinkowskiEngine [26].

***Testing Datasets.*** For testing, we used two datasets:
- 8i Voxelized Full Bodies (8iVFB) [27] : includes four sequences- *longdress*, *loot*, *redandblack*, and *soldier*.
- Owlii dynamic human mesh (Owlii) [28] : includes two sequences- *basketball_player*, and *dancer*.

All the training and testing experiments were conducted on a system equipped with an Intel i7-14700 CPU and an Nvidia 4090 RTX GPU.

***Training setting Details.*** The network model was optimized using the Adam optimizer, with $\beta_1$ and $\beta_2$ set to 0.9 and 0.999, respectively. The learning rate is initialized at 8e-5 and halved every 20 epochs until it decreases to 2e-5. The model was randomly initialized and trained for a maximum of 60 epochs. Stage 1 and stage 2 were dedicated to training for high-rate scenarios, while stage 3 refined all components to enhance overall performance across various configurations. The loss function parameters were set as follows. In (1), $\lambda_A$ was set to 0.03 and $\alpha$ was set to 2. In (2), $\lambda_G$ was set to 6 and $\lambda_{MSE}$ was set to 1.5. In (3), $\lambda_A$ was set to 0.03, 0.04, 0.04, 0.05, 0.05, and 0.05, while $\lambda_G$ was set to 6, 4, 4, 8, 12, and 20. Additionally, $\lambda_t$ in (3) is set to 0.5, 0.25, 0.125, 0.05, 0.015, and 0.005 to obtain different models.

***Evaluation Metric.*** PCC traditionally relies on metrics such as D1-PSNR and D2-PSNR for geometry quality, and Y-PSNR and YUV-PSNR for attribute evaluation. However, each of these metrics alone is insufficient for assessing overall quality. Similar to JPEG Pleno PCC, in additional to D1-PSNR and Y-PSNR, we employ PCQM [29] as a unified evaluation metric. Findings from [30] indicate that prioritizing attribute bitrate often results in slightly better perceptual quality. This insight guides the optimal allocation of bitrates when jointly compressing geometry and attributes.

Table 2. Complexity analysis for encoding and decoding sequence "Soldier"

| | YOGA [16] | | JPEG Pleno PCC [19] Low rate (with SR) | | JPEG Pleno PCC [19] High rate (without SR) | | DeepPCC [17] | | Proposed SEDD-PCC | |
|---|---|---|---|---|---|---|---|---|---|---|
| Model Size (MB) | Geo. + Att. 169.5 MB | | Geo. + Att. 87.5 MB | | Geo. + Att. 64.8 MB | | Geo. + Att. 124.3 MB | | Geo. + Att. 32.6 MB | |
| Enc/Dec (s) | Enc | Dec | Enc | Dec | Enc | Dec | Enc | Dec | Enc | Dec |
| | --- | | 53.8 | 46.2 | 47.2 | 34.4 | --- | | 0.43 | 0.77 |

## 3.2 Performance Evaluation

To comprehensively evaluate the performance of our SEDD-PCC, we compared it against standard MPEG benchmarks, including G-PCC TMC13 v23 Octree-RAHT and V-PCC TMC2 v22, following the MPEG Common Test Condition (CTC) [31]. Additionally, we compare our approach with several learning-based techniques for joint point cloud compression, specifically YOGA [16], DeepPCC [17], JPEG Pleno [19], and Unicorn [21]. The R-D performance in terms of 1-PCQM is illustrated in Fig. 4, where SEDD-PCC consistently outperforms G-PCC. For a more comprehensive comparison, Table 1 presents the BD-BR (%) of our method, as well as various benchmarks when G-PCC is served as the anchor. Notably, compared to G-PCC, our approach achieves substantial bitrate reductions, with an average saving of 75.0% in D1-PSNR, 32.6% in Y-PSNR, and 33.2% in 1−PCQM. These results highlight the superior coding performance of SEDD-PCC, marking a significant advancement in joint point cloud compression.

## 3.3 Complexity Evaluation

Table 2 presents the complexity analysis in terms of model size and encoding/decoding time. Due to availability constraints, only a subset of learning-based joint coding schemes is included. Our SEDD-PCC has a model size of 32.6 MB, making it significantly lighter than other methods. Additionally, it eliminates the need for recoloring, reducing the whole processing time. Eventually, its encoding and decoding process is exceptionally fast.

## 3.4 Ablation Study

### 3.4.1 Knowledge Distillation

In Stage 2 of training, we employ a teacher model to guide the initialization of the geometry decoder and the transform module through knowledge distillation. This process minimizes feature discrepancies between the teacher and student models, enhancing the geometry decoder's performance.

To quantify the benefits of this approach, Table 3 presents the BD-rate improvements when incorporating a teacher model in Training Stage 2. The results indicate a 6.9% bitrate reduction in D1-PSNR, along with a 3.8% reduction in Y-PSNR and a 3.7% decrease in 1-PCQM. These findings confirm that knowledge distillation enhances the overall training effectiveness and contributes to improved compression efficiency.

Table 3. Ablation study regarding the adoption of Teacher Model and Transform Module

| Component | Average BD-BR(%) | | |
|---|---|---|---|
| | D1-PSNR | Y-PSNR | 1-PCQM |
| Teacher Model | -6.9 | -3.8 | -3.7 |
| Transform Module | -7.1 | -6.5 | -3.9 |

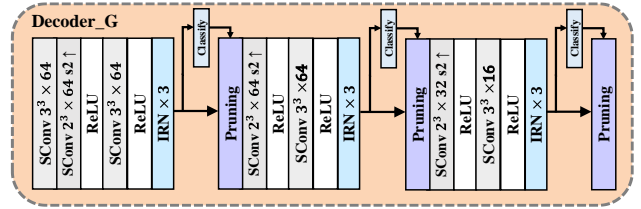

Fig. 5. The network of the alternative geometry decoder when transform module is removed.

### 3.4.2 Transform Module

The transform module is designed to adapt features from the shared encoder into geometry-specific features. To evaluate its necessity, we explore an alternative geometry decoder that does not rely on a transform module, as illustrated in Fig. 5.

Table 3 presents the BD-rate results comparing SEDD-PCC with this alternative design that omits the transform module in the geometry decoder. Without the transform module, the bitrate is reduced by 7.1% for D1-PSNR, 6.5% for Y-PSNR, and 3.9% for 1-PCQM. The results demonstrate that incorporating the transform module improves encoding performance, highlighting its critical role in effectively utilizing shared encoder features for geometry reconstruction.

## 4. CONCLUSION

In this study, we present SEDD-PCC, a unified framework that employs a single encoder and two decoders to jointly compress and reconstruct both the geometry and attributes of point clouds. By eliminating the recoloring process, our approach enables a fully end-to-end learned framework, making it more efficient than existing unified methods that rely on recoloring. With a simple yet effective network design, SEDD-PCC achieves highly competitive compression performance while maintaining efficiency. For future work, improving attribute compression through more advanced feature extraction techniques could further enhance quality. Additionally, developing a more efficient entropy model remains a key focus. In our proposed joint compression approach, bit allocation is determined by the loss function, which may not be optimal. Designing a loss function that better balances geometry and attributes remains an important direction for further research.

# REFERENCES

[1] V-PCC codec description, in ISO/IEC JTC 1/SC 29/WG 7 N00100, 2021.

[2] G-PCC codec description, in ISO/IEC JTC 1/SC 29/WG 7 N00271, 2022.

[3] J. Wang, D. Ding, Z. Li and Z. Ma, "Multiscale point cloud geometry compression," *Proc. of Data Compression Conference* (DCC), 2021, pp. 73-82.

[4] J. Wang, D. Ding, Z. Li, X. Feng, C. Cao and Z. Ma, "Sparse tensor-based multiscale representation for point cloud geometry compression," *IEEE Transactions on Pattern Analysis and Machine Intelligence*, vol. 45, no. 7, pp. 9055-9071, 1 July 2023.

[5] J. -C. Chiang, J, -J. Chiu, and M. Yim, "ANFPCGC++: Point cloud geometry coding using augmented normalizing flows and Transformer-based entropy model," *IEEE Access*, 2024.

[6] J. Pang, M. A. Lodhi, and D. Tian, "GRASP-Net: Geometric residual analysis and synthesis for point cloud compression," *ACM MM*, 2022.

[7] G. Liu, J. Wang, D. Ding and Z. Ma, "PCGFormer: Lossy point cloud geometry compression via local self-attention," *Proc. of IEEE International Conference on Visual Communications and Image Processing* (VCIP), 2022.

[8] X. Sheng, L. Li, D. Liu, Z. Xiong, Z. Li and F. Wu, "Deep-PCAC: An end-to-end deep lossy compression framework for point cloud attributes," *IEEE Transactions on Multimedia*, vol. 24, pp. 2617-2632, 2022.

[9] J. Wang and Z. Ma, "Sparse tensor-based point cloud attribute compression," *Proc. of IEEE International Conference on Multimedia Information Processing and Retrieval*, 2022, pp. 59-64.

[10] R. B. Pinheiro, J. -E. Marvie, G. Valenzise and F. Dufaux, "Reducing the complexity of normalizing flow architectures for point cloud attribute compression," *Proc. of IEEE International Conference on Acoustics, Speech and Signal Processing* (ICASSP), 2024.

[11] T. -P. Lin, M. Yim, J. -C. Chiang, W. -H. Peng and W. -N. Lie, "Sparse tensor-based point cloud attribute compression using augmented normalizing flows," *Proc. of Asia Pacific Signal and Information Processing Association Annual Summit and Conference* (APSIPA ASC), 2023, pp. 1739-1744.

[12] X. Mao, H. Yuan, T. Guo, S. Jiang, R. Hamzaoui and S. Kwong, "SPAC: Sampling-based Progressive Attribute Compression for Dense Point Clouds," *arXiv preprint arXiv:2409.10293*, 2024.

[13] S. Umair, B. Kathariya, Z. Li, A. Akhtar, G. Van der Auwera, "ResNeRF-PCAC: Super Resolving Residual Learning NeRF for High Efficiency Point Cloud Attributes Coding," *Proc. of IEEE International Conference on Image Processing* (ICIP), 2024, pp. 3540-3546.

[14] E. Alexiou, K. Tung, and T. Ebrahimi, "Towards neural network approaches for point cloud compression," *Proc. of Applications of digital image processing XLIII*, vol. 11510, pp. 18-37, 2020.

[15] A. F. R. Guarda, M. Ruivo, L. Coelho, A. Seleem, N. M. M. Rodrigues and F. Pereira, "Deep Learning-Based Point Cloud Coding and Super-Resolution: a Joint Geometry and Color approach," *IEEE Trans. Multimedia*, doi: 10.1109/TMM.2023.3338081.

[16] J. Zhang, T. Chen, D. Ding, and Z. Ma, "YOGA: Yet another geometry-based point cloud compressor," *Proc. of ACM International Conference on Multimedia*, 2023, pp. 9070-9081.

[17] J. Zhang, G. Liu, J. Zhang, D. Ding, and Z. Ma, "DeepPCC: Learned Lossy Point Cloud Compression, " *accepted by IEEE Trans. Emerging Topics in Computational Intelligence*, Sept. 2024.

[18] "Verification Model Description for JPEG Pleno Learning-based Point Cloud Coding v4.0," in ISO/IEC JTC1/SC29/WG1 N100709, 102nd Meeting, San Francisco, CA, USA, Jan. 2024.

[19] A. F. Guarda, N. M. Rodrigues, and F. Pereira, "The JPEG Pleno Learning-based Point Cloud Coding Standard: Serving Man and Machine," *arXiv preprint arXiv:2409.08130*, 2024.

[20] J. Wang, R. Xue, J. Li, D. Ding, Y. Lin, and Z. Ma, "A Versatile Point Cloud Compressor Using Universal Multiscale Conditional Coding - Part I: Geometry," *IEEE Trans. Pattern Analysis and Machine Intelligence*, 47(1):269-287, Jan. 2025.

[21] J. Wang, R. Xue, J. Li, D. Ding, Y. Lin, and Z. Ma, "A Versatile Point Cloud Compressor Using Universal Multiscale Conditional Coding - Part II: Attribute," *IEEE Trans. Pattern Analysis and Machine Intelligence*, 47(1):252-268, Jan. 2025.

[22] J. Ascenso, E. Alshina and T. Ebrahimi, "The JPEG AI standard: providing efficient human and machine visual data consumption," *IEEE MultiMedia*, vol. 30, no. 1, pp. 100-111, 1 Jan.-Mar. 2023, doi: 10.1109/MMUL.2023.3245919.

[23] C. Szegedy, S. Ioffe, V. Vanhoucke and A. Alemi, "Inception-v4, inception-resnet and the impact of residual connections on learning," In *Proceedings of the AAAI conference on artificial intelligence* (Vol. 31, No. 1).

[24] "SparsePCGCv1 Update: Improvements on dense/sparse/LiDAR point clouds," ISO/IEC JTC 1/SC 29/WG 7 m60352, Online, July. 2022.

[25] A. Dai, A. X. Chang, M. Savva, M. Halber, T. Funkhouser and M. Nießner, "ScanNet: richly-annotated 3d reconstructions of indoor scenes," *IEEE Conference on Computer Vision and Pattern Recognition*, 2017, pp. 2432-2443.

[26] C. Choy, J. Gwak and S. Savarese, "4D spatio-temporal ConvNets: Minkowski convolutional neural networks," *Proc. of IEEE/CVF Conference on Computer Vision and Pattern Recognition (CVPR)*, 2019, pp. 3070-3079.

[27] E. d'Eon, B. Harrison, T. Myers, and P. A. Chou, "8I voxelized full bodies - a voxelized point cloud dataset," ISO/IEC JTC1/SC29 Joint WG11/WG1 (MPEG/JPEG) m40059/M74006, Jan. 2017.

[28] X. Yi, L. Yao, and W. Ziyu, " Owlii Dynamic Human Mesh Sequence Dataset, ISO/IEC Standard JTC1/SC29/WG11 m41658, ISO/IEC: Washington, DC, USA, Oct. 2017.

[29] G. Meynet, Ya. Nehmé, J. Digne, and G. Lavoué. "PCQM: A full-reference quality metric for colored 3D point clouds," *Proc. of International Conference on Quality of Multimedia Experience* (QoMEX), 2020.

[30] J. Prazeres, R. Rodrigues, M. Pereira and A. M. G. Pinheiro, "Quality Analysis of the Coding Bitrate Tradeoff Between Geometry and Attributes for Colored Point Clouds. " *arXiv preprint arXiv:2410.21613*, 2024.

[31] "Common test conditions for point cloud compression," in ISO/IEC JTC1/SC29/WG11 MPEG output document N19084.